\documentclass{article}

    \PassOptionsToPackage{numbers, compress}{natbib}

\usepackage[preprint]{neurips_2024}

\usepackage[utf8]{inputenc} 
\usepackage[T1]{fontenc}    
\usepackage{hyperref}       
\usepackage{url}            
\usepackage{booktabs}       
\usepackage{amsfonts}       
\usepackage{nicefrac}       
\usepackage{microtype}      
\usepackage{xcolor}         
\usepackage{bm}
\usepackage{amsmath}
\usepackage{bbding}
\usepackage{algorithm}
\usepackage{algorithmic}
\usepackage{graphicx}
\usepackage{multirow}
\usepackage{subfig}
\usepackage{wrapfig}

\title{Hashing for Protein Structure Similarity Search}

\author{%
 Jin~Han, Wu-Jun~Li\thanks{Corresponding author} \\
National Key Laboratory for Novel Software Technology\\
  Department of Computer Science and Technology\\
  Nanjing University, Nanjing 210023, China\\
  \texttt{hanjin@smail.nju.edu.cn,liwujun@nju.edu.cn} \\
}

\begin{document}

\maketitle

\begin{abstract}
Protein structure similarity search~(PSSS), which tries to search proteins with similar structures, plays a crucial role across diverse domains from drug design to protein function prediction and molecular evolution.
Traditional alignment-based PSSS methods, which directly calculate alignment on the protein structures, are highly time-consuming with high memory cost.
Recently, alignment-free methods, which represent protein structures as fixed-length real-valued vectors, are proposed for PSSS. Although these methods have lower time and memory cost than alignment-based methods, their time and memory cost is still too high for large-scale PSSS, and their accuracy is unsatisfactory.
In this paper, we propose a novel method, called $\underline{\text{p}}$r$\underline{\text{o}}$tein $\underline{\text{s}}$tructure $\underline{\text{h}}$ashing~(POSH), for PSSS.
POSH learns a binary vector representation for each protein structure, which can dramatically reduce the time and memory cost for PSSS compared with real-valued vector representation based methods. 
Furthermore, in POSH we also propose expressive hand-crafted features and a structure encoder to well model both node and edge interactions in proteins.
Experimental results on real datasets show that POSH can outperform other methods to achieve state-of-the-art accuracy.
Furthermore, POSH achieves a memory saving of more than six times and speed improvement of more than four times, compared with other methods.
\end{abstract}

\section{Introduction}
Proteins play essential roles in biological systems by binding with various ligands. Proteins with similar structures often share similar functions.
Hence, protein structure similarity search~(PSSS), which tries to search proteins with similar structures, plays a crucial role across diverse domains from drug design to protein function prediction and molecular evolution.

Traditional PSSS methods, which are often called alignment-based methods, directly calculate alignment on the protein structures. Representative alignment-based methods include TM-align~\cite{zhang2005tm}, MATT~\cite{menke2008matt} and several others~\cite{shindyalov1998protein,ye2003flexible,wang2013protein}.
Since identifying an optimal alignment between a pair of structures is an NP-hard problem~\cite{lathrop1994protein}, alignment-based methods are typically highly time-consuming with high memory cost even if heuristic techniques are adopted in these methods. Taking TM-align as an example, aligning a protein structure with 200 amino acids against the SCOPe database~\cite{DBLP:journals/nar/FoxBC14} with 14323 protein structures takes approximately 30 minutes. Note that this is the time cost for only a single query. The memory cost for storing the SCOPe database is approximately 4GB. With the development of Cryo-EM and protein structure prediction techniques, such as \mbox{Alphafold 2}~\cite{jumper2021highly}, RoseTTAFold~\cite{baek2021accurate} and ESMFold~\cite{lin2022language}, the number of proteins with known structures grows rapidly. For example, \mbox{Alphafold 2} has predicted structures with high confidence for over 200 million proteins, which has been used to construct the Alphafold database~\cite{varadi2022alphafold}. On large-scale datasets like Alphafold database, alignment-based methods will have huge time and memory cost, even become infeasible. 

To reduce the cost for PSSS, alignment-free methods, which represent protein structures as fixed-length real-valued vectors, are proposed. Existing alignment-free methods can be categorized into non-learning methods and learning-based methods. Non-learning methods~\cite{rogen2003automatic,zotenko2006secondary} generate vector representations based on hand-crafted features. Learning-based methods~\cite{liu2018learning,xia2022fast} generate vector representations by learning from data. PSSS is performed by calculating and ranking the similarity between the vector of the query and those of the database. With vector representation, alignment-free methods have
lower time and memory cost than alignment-based methods. However, it is still time-consuming to calculate the similarity between real-valued vectors, and the memory cost for storing real-valued vectors is still too high for large-scale datasets.
Furthermore, the accuracy of existing alignment-free methods is unsatisfactory because these methods have limitation in representing the complex protein structure.
More specifically, the hand-crafted features in non-learning methods fail to model the complex and irregular three-dimensional protein shape.
Existing learning-based methods suffer from the lack of expressive features and cannot model edge interactions in proteins.

In this paper, we propose a novel method, called \underline{p}r\underline{o}tein \underline{s}tructure \underline{h}ashing~(POSH), for PSSS. 
The main contributions of POSH are outlined as follows:
\begin{itemize} 
    \item To the best of our knowledge, POSH is the first hashing method for PSSS.   
   \item POSH learns a binary vector~(or called hash code) representation for each protein structure, which can dramatically reduce the time and memory cost for PSSS compared with real-valued vector representation based methods. 
    \item Furthermore, in POSH we also propose expressive hand-crafted features and a structure encoder to well model both node and edge interactions in proteins.
    \item Experimental results on real datasets show that POSH can outperform other methods to achieve state-of-the-art accuracy.
    Furthermore, POSH achieves a memory saving of more than six times and speed improvement of more than four times, compared with other methods.
\end{itemize}

\section{Related Works}
\paragraph{PSSS}  
Given a query protein, PSSS tries to search~(return) similar proteins from a protein database, where the similarity between two proteins is computed based on the three-dimensional structures of these two proteins. Existing PSSS methods can be categorized into alignment-based methods and alignment-free methods. 

Template modeling score~(TM-score)~\cite{zhang2004scoring} is a well-known metric for assessing the topological similarity of protein structures, which has been adopted in the evaluation of Critical Assessment of protein Structure Prediction (CASP) ~\cite{jones2001critically}. 
TM-align~\cite{zhang2005tm} is an alignment-based method using TM-score.
TM-align employs heuristic dynamic programming~(DP) to generate residue-to-residue alignments, but the process of performing DP on the similarity score matrix is time-consuming.
There have also appeared other alignment-based methods like ~\cite{shindyalov1998protein,ye2003flexible,menke2008matt,wang2013protein}. However, none of them can avoid the time-consuming procedure of aligning the coordinates of individual atoms between structures.
Furthermore, since these alignment-based methods directly perform alignment on the raw protein structures, the memory cost for storing the raw protein structures is also high.

Several alignment-free methods have been proposed to overcome the inefficiency of alignment-based methods by representing protein structures as fixed-length real-valued vectors, which can be further categorized into non-learning methods and learning-based methods. 
Non-learning methods use hand-crafted features to represent the protein structures.
For example, SGM~\cite{rogen2003automatic} represents protein structures using 30 structural measures, and SSEF~\cite{zotenko2006secondary} utilizes secondary structure information to map structures into vectors. 
The shortcoming of non-learning methods is that the hand-crafted features fail to model the complex and irregular three-dimensional protein shape.
Hence, learning-based methods are proposed to learn more informative representations from the hand-crafted features.
DeepFold~\cite{liu2018learning} and GraSR~\cite{xia2022fast} are two representative learning-based methods.  
DeepFold extracts a distance map from the protein structure and employs convolutional neural network~(CNN) to learn a vector representation, but it neglects the graph property of protein structures.
GraSR~\cite{xia2022fast} models the protein structure as a graph with node features and utilizes long short-term memory~(LSTM)~\cite{shi2015convolutional} and graph convolutional network~(GCN)~\cite{kipf2016semi} to update the node features. 
However, GraSR only considers the relations of local neighboring nodes and fails to model the edge interactions in proteins. Furthermore, all existing alignment-free methods adopt real-valued vectors for feature representation, which still have high time and memory cost for large-scale datasets.  

\paragraph{Hashing}
Hashing has been widely used for efficient search in many areas~\cite{Li2016feature,cao2017hashnet,xu2020hashing,ikuya2021efficeint,wang2021learning}. 
Hashing uses a hash function to map each data sample to a binary vector~(or called hash code) while preserving the similarity in the original space. The hash function can be manually designed or learned from training data. In this paper, we focus on learned hash functions because learned functions can typically achieve better performance. Binary hash codes have advantages in search speed and memory cost. To the best of our knowledge, no research has explored the application of hashing for PSSS.

\section{Method}
\begin{figure*}[t]
\centering
\includegraphics[scale=0.46]{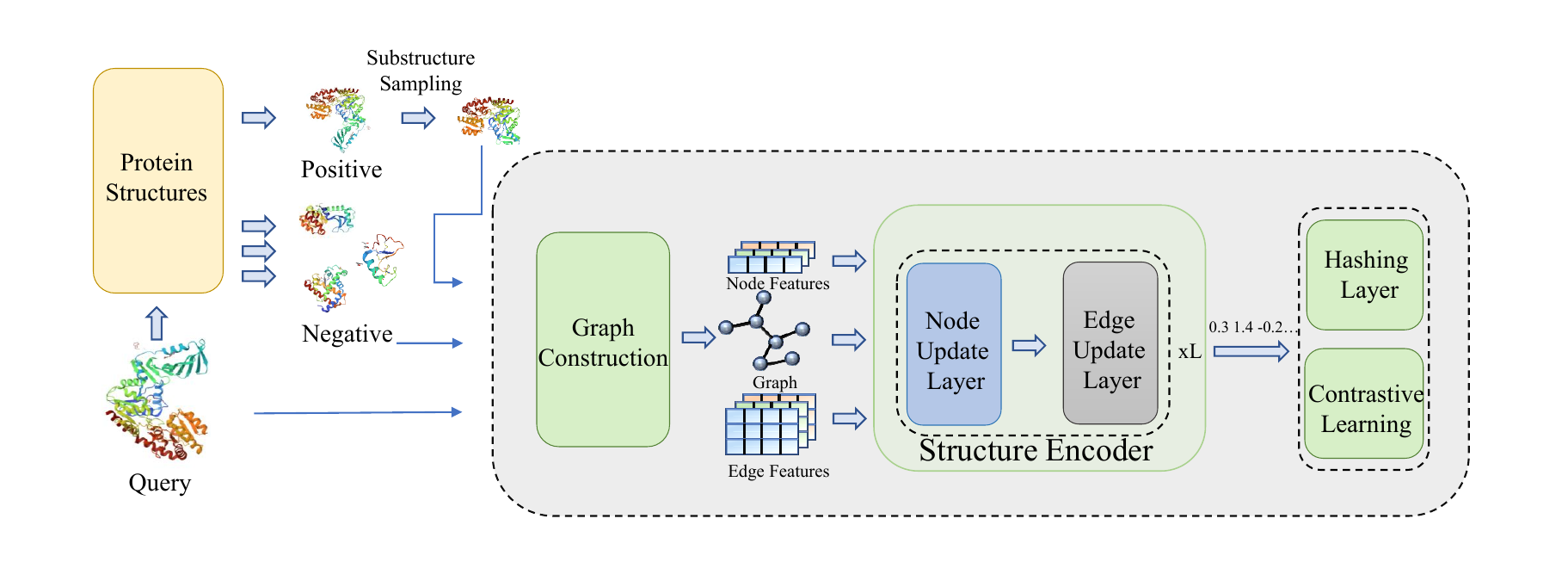}
\caption{The architecture of POSH}
\label{figure1}
\end{figure*}
Figure~\ref{figure1} illustrates the architecture of our proposed POSH for PSSS.
The architecture comprises several key components: graph construction, structure encoder, hashing layer and contrastive learning component for objective function, and a data sampling strategy to sample negative samples and substructures of positive samples.

\subsection{Graph Construction}\label{sec:graphConstruction}
We construct a graph to represent each protein, and introduce several techniques to extract features for both nodes and edges from the raw protein structure. 

\paragraph{Graph Structure} 
A protein is represented as a graph $\mathcal{G(V, E)}$ with $\mathcal{V}$ denoting the set of nodes and $\mathcal{E}$ denoting the set of edges. 
Each amino acid corresponds to a node in the graph, and the edges are constructed by connecting each node with its $k$ nearest neighbors~($k$NN). 
The distance between nodes is measured by the Euclidean distance between the raw node features of the $C_{\alpha}$ atoms. The raw node features for distance computation are introduced in the following content.

\begin{figure}[t]
\centering
\includegraphics[scale=0.8]{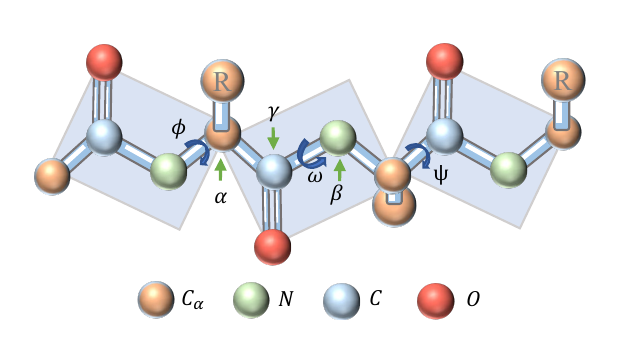}
\caption{Illustration of bond angles and dihedral angles. The letter R denotes the side chain of the amino acid.}
\label{figure2}
\end{figure}

\paragraph{Raw Node Features} 
Although the TM-score of proteins is typically computed based on the coordinates of $C_\alpha$ atoms, we also incorporate other backbone atoms to construct node features.
Because $C_\alpha$ atoms and the other backbone atoms are constrained by the peptide plane and form the 3D protein shape together, it is reasonable to include them in node features. In particular, we calculate the bond and dihedral angles for each residual as our node features. 
As shown in Figure \ref{figure2}, the bond angle is the angle formed between two covalent bonds that share a common atom, and the dihedral angles, also known as torsion angles, refer to the angles between planes defined by four consecutive atoms in the protein backbone. 
For residual $i$, we denote the bond angles of $N_i\mbox{-}C_{\alpha_i}\mbox{-}C_i, C_{i-1}\mbox{-}N_i\mbox{-}C_{\alpha_i}, C_{\alpha_i}\mbox{-}C_i\mbox{-}N_{i+1}$ as $\alpha_i,\beta_i,\gamma_i$, and the dihedral angles of $N_{i-1}\mbox{-}C_{\alpha_{i-1}}, C_{\alpha_{i-1}}\mbox{-}C_{i-1},C_i\mbox{-}N_i$ as $\phi_i,\psi_i,\omega_i$. 
The final node features are computed by $\{sin,cos\}\circ\{\alpha_i,\beta_i, \gamma_i, \phi_i,\psi_i,\omega_i\}$. 
We use $\bm{V}\in R^{n\times d_v}$ to denote these hand-crafted node features~(also called raw node features), where $n$ is the number of nodes and $d_v$ is the dimension of the raw node features. 
\paragraph{Raw Edge Features}  
Unlike existing learning-based methods which do not use edge features, we additionally extract edge features for better description of the structure. We consider the following five common types of atoms in the protein chain: $C$, $C_\alpha$, $N$, $O$, and $C_\beta$. 
If a $C_\beta$ atom is missing, we employ a technique in~\cite{dauparas2022robust} to add a virtual $C_\beta$ atom based on the constraint of the other backbone atoms. 
To calculate the distance between the $i$th node and the $j$th node, we compute the Euclidean distance $\Vert X_i-Y_j\Vert$, where $X_i$ is chosen from the set $\{C_i, C_{\alpha_i}, N_i, O_i, C_{\beta_i}\}$, and $Y_j$ is chosen from the set $\{C_j, C_{\alpha_j}, N_j, O_j, C_{\beta_j}\}$. 
Here, $X_i$ and $Y_j$ represent the respective coordinates of the atoms. 
Subsequently, we encode the distances using a Gaussian radial basis function~(RBF). 
The final representation of the edge features is computed by $\text{RBF}_k(\Vert X_i-Y_j\Vert)$, where $\text{RBF}_k$ denotes the $k$th RBF employed in the encoding process. 
We use $\bm{E} \in R^{m\times d_e}$ to denote these hand-crafted edge features~(also called raw edge features), where $m$ is the number of edges and $d_e$ is the dimension of the raw edge features.

\subsection{Structure Encoder} 
We propose a novel structure encoder, which is a deep graph neural network with $L$ layers, to learn more informative node features from the hand-crafted raw features constructed in Section~\ref{sec:graphConstruction}.
The structure encoder consists of node update layer and edge update layer.
We first map the hand-crafted raw features of node $i$ (the $i$th row of $\bm{V}$) and the hand-crafted raw features of edge $k$~(the $k$th row of $\bm{E}$) to have the same hidden dimension with a linear mapping, to obtain the initial node representation $\bm{h}_i^0$ and edge representation $\bm{e}_{ij}^0$, respectively. Here, edge $k$ is from node $i$ to node $j$, and its initial representation is $\bm{e}_{ij}^0$.
\begin{figure*}[t]
\centering
\includegraphics[scale=0.65]{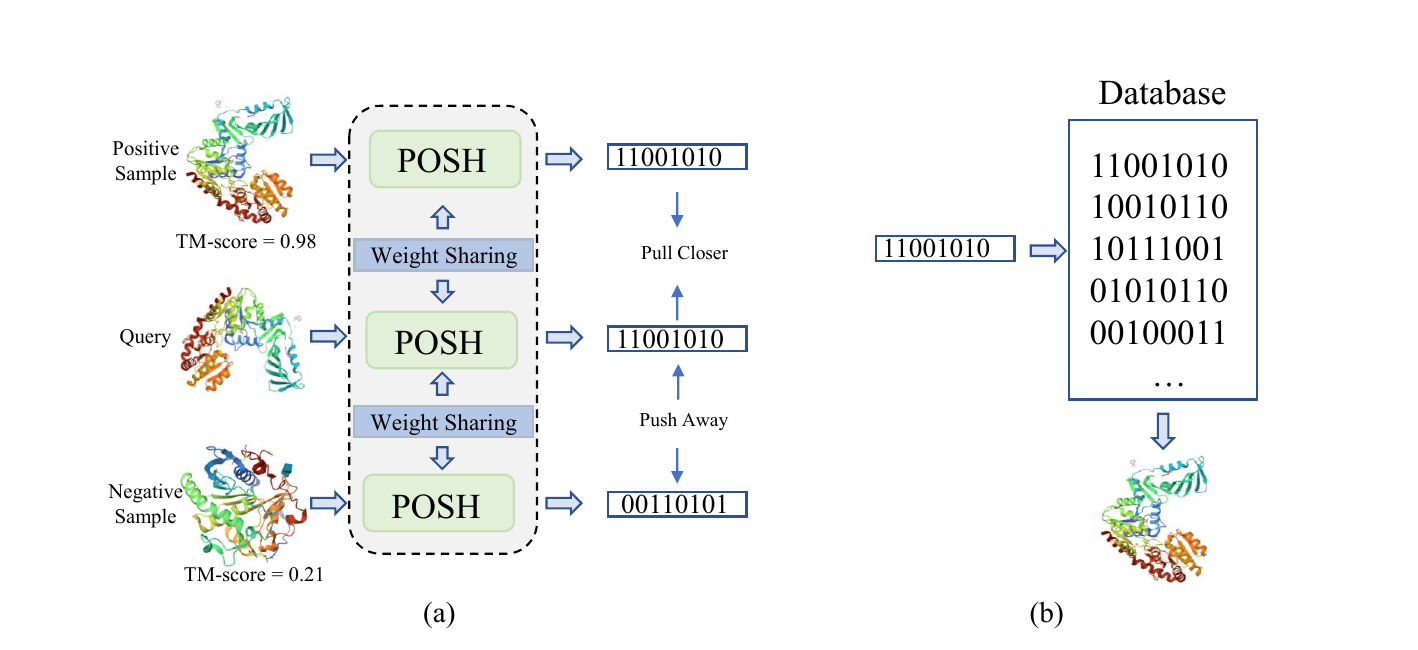}
\caption{POSH for protein structure similarity search. 
(a) In the training phase, the distance between binary hash codes will be minimized or maximized depending on the original similarity of the samples. 
(b) In the testing phase, the binary hash code for the new-coming query protein structure can be obtained from POSH, and then be used to search the database to get similar protein structures.}
\label{figure3}
\end{figure*}
\paragraph{Node Update Layer} 
For each node $i$ in layer $l$, we update its representation $\bm{h}_i^l$ by aggregating features from its neighboring nodes and adjacent edges.
This message passing mechanism is defined as follows:
\begin{equation}
    \begin{aligned}
    \bm{u}_i^{l}&=\bm{h}_i^{l}+\frac{1}{|\mathcal{N}_i|}\sum_{j\in \mathcal{N}_i}NodeMLP(\bm{h}_i^l\Vert \bm{h}_j^l\Vert \bm{e}_{ij}^l),\\
    \tilde{\bm{h}}_i^{l}&=\Phi(\bm{h}_i^{l}+MLP(\bm{u}^{l}_i)),
    \end{aligned}
\end{equation}
where $\mathcal{N}_i$ denotes the index set of neighboring nodes of node $i$ and $|\mathcal{N}_i|$ denotes the number of nodes in $\mathcal{N}_i$. $\Vert$ denotes the concatenation operation. 
$NodeMLP$ is a two-layer feed-forward neural network.
The updated node representation $\bm{u}_i^{l}$ is further updated by a $MLP$, which also denotes a two-layer feed-forward neural network and a residual connection from $\bm{h}_i^{l}$ is then added.
$\Phi$ is a follow-up batchnorm function.
This straightforward message passing mechanism is memory efficient so that we can sample more negative samples in a single mini-batch, the detail of which will be introduced in Section~\ref{sec:dataSampling}.
\paragraph{Edge Update Layer} 
As proved in~\cite{gao2022pifold,DBLP:conf/iclr/ZhangXJCLD023}, neglecting the update of edge features could lead to suboptimal performance. Hence, we also perform edge message passing in our method.
For each edge $(i,j)$, we aggregate features from the connected node $i$ and node $j$, the result of which is then added to the original edge representation. 
Our edge message passing mechanism is defined as follows:
\begin{equation}
\begin{aligned}
\bm{e}_{ij}^{l+1}=\Phi(\bm{e}_{ij}^{l}+EdgeMLP(\tilde{\bm{h}}_i^l\Vert \tilde{\bm{h}}_j^l\Vert \bm{e}_{ij}^l)).\\
\end{aligned}
\end{equation}
Here, the updated node representation $\tilde{\bm{h}}_i^{l}$ and $\tilde{\bm{h}}_j^{l}$ are concatenated with the edge representation $\bm{e}_{ij}^{l}$. The concatenated result will then be updated with an $EdgeMLP$, which is also a two-layer feed-forward neural network.
For the next layer, we set $\bm{h}_i^{l+1}=\tilde{\bm{h}}_i^{l}$.

\subsection{Objective Function}
The output of the structure encoder is denoted as $\bm{h}_i^L$, where $L$ is the number of stacked layers.
For each protein structure $t$, the representation of all the nodes in protein structure $t$ is $\{\bm{h}_i^L\}_{i\in \mathcal{V}_{t}}$.The final representation $\bm{y}_{t}$ of protein structure $t$ is computed by $\bm{y}_{t} = Linear(f_p(\{\bm{h}_i^L\}_{i\in \mathcal{V}_{t}}))$,
where $f_p$ is a max pooling function, and $Linear$ is a linear mapping.
Note that the resulting vector $\bm{y}_t$ is real-valued at this stage. 

Let $\bm{b}_t\in\{-1,1\}^d$ denote the binary hash code\footnote{During training, we use $\bm{b}_t\in\{-1,1\}^d$ to denote the hash code for the convenience of modeling. After training, we change $-1$ to 0 to get the final hash code representation.} for representing protein structure $t$. We can get $\bm{b}_t=sign(\bm{y}_t)$. To enable end-to-end learning of the binary hash code, we add a hashing layer to the model. More specifically, 
we introduce a loss to encourage $\bm{y}_t$ to approach the binary hash code $\bm{b}_t$ as in~\cite{Li2016feature}: 
 \begin{equation}
    \mathcal{L}_\text{hash} = \sum_{t}\Vert \bm{y}_t-\bm{b}_t \Vert_2^2 + \gamma\sum_{t} (\bm{y}_t\bm{1})^2,
\end{equation}
where $\gamma$ is a hyperparameter and the term $(\bm{y}_t\bm{1} )^2$ is designed to avoid bias or skewed representations of protein structures.

Each time, we sample protein structures containing one protein structure as query, one positive sample, and $K$ negative samples. Here, positive samples are those labeled to be similar to the query, and negative samples are those labeled to be dissimilar to the query.
We denote the query sample as $Q$, the positive sample as $P$, and the negative samples as $F_1, F_2, \cdots, F_K$.
As depicted in Figure \ref{figure3}~(a), our objective is to minimize distance for similar samples and maximize distance for dissimilar ones. 
To achieve this, we employ the InfoNCE loss \cite{oord2018representation} from contrastive learning,  which aims to minimize the negative log-likelihood of the similar pairs. 
To ensure stability during training and avoid gradient exploding caused by the exponential function, we apply $L_2$ normalization to each $\bm{y}_t$ and obtain $\hat{\bm{y}}_t=\text{norm}(\bm{y}_t)$. Then we have:
\begin{equation}
    \mathcal{L}_\text{sim}=-\log(\frac{exp({\hat{\bm{y}}_Q\cdot \hat{\bm{y}}_{P}/\tau})}{exp({\hat{\bm{y}}_Q\cdot \hat{\bm{y}}_{P}/\tau})+\sum_{i=1}^Kexp({\hat{\bm{y}}_Q\cdot \hat{\bm{y}}_{F_i}/\tau})}),
\end{equation}
where $\tau$ is a temperature hyper-parameter that controls the relative distance between samples, the denominator represents the sum over one positive sample and $K$ negative samples. 
The similarity of two vectors is computed using the dot product ($\cdot$). 

The entire model can  be trained in an end-to-end manner, utilizing the following combined objective function:
\begin{equation}
    \mathcal{L}=\mathcal{L}_\text{sim}+\lambda\mathcal{L}_\text{hash},
\end{equation}
where $\lambda$ is a hyperparameter for balancing the two loss functions.

In the testing phase, a protein structure will be fed into POSH, and first get a real-valued representation $\bm{y}_t$. 
Then this real-valued representation undergoes a $sign(\cdot)$ function to get a binary hash code representation.
As depicted in Figure \ref{figure3}~(b), the target database to be searched will be pre-encoded.
Each new-coming query is first encoded as a binary vector~(hash code) and subsequently the hash code is used to search in the target database based on Hamming distance.
Due to the discrete nature of the Hamming distance, it is possible to have multiple samples in the database with the same Hamming distance to the query.
We apply an additional scaling factor to the distance to distinguish the samples with the same Hamming distance, which is defined as follows:
\begin{equation}
    dist'=dist/(1+\frac{|l_q-l_t|}{l_{max}}),
\end{equation}
where $dist$ is the original Hamming distance between the query protein structure and a protein structure in database, $l_q$ is the length of the query protein structure, $l_t$ is the length of the protein structure in database, and $l_{max}$ is the maximum length of the protein structure in database.
This scaling factor imposes penalty on pairs with significantly different lengths.

\subsection{Data Sampling Strategy}\label{sec:dataSampling}
In this subsection, we introduce how to sample positive and negative samples according to the query sample, and a substructure sampling strategy is further proposed to enhance the diversity of positive samples.
\paragraph{Training Data Sampling} 
We first calculate the pairwise TM-score~\cite{zhang2004scoring} for protein structures using TM-align~\cite{zhang2005tm}, serving as the ground-truth measurement of structural similarity. 
In the training phase, we define two protein structures, denoted as $(P_a, P_b)$, to be similar based on the following criterion: 
if the TM-score between $P_a$ and $P_b$ is larger than a threshold $\rho$ multiplied by the maximum TM-score between $P_a$ and any other structure $P_i$ in the training set $\mathcal{D}$ excluding $P_a$, as shown below:
\begin{equation}
\label{eq1}
    TM(P_a, P_b) \geq \rho\cdot max(\{TM(P_a,P_i) |{P_i\in\mathcal{D}\setminus P_a}\}),
\end{equation}
where $TM(\cdot)$ denotes the TM-score, $\rho$ is a hyperparameter in (0, 1).
In each time, we sample a mini-batch of data comprising one protein structure as query, one positive sample, and $K$ negative samples. 
The positive sample means structurally similar to the query, while the negative samples are dissimilar.

\paragraph{Substructure Sampling} 
The above sampling strategy might cause an issue. 
We observe that some protein structures have few (less than 5) similar protein structures~(positive samples), and hence, the same structure will be repeatedly sampled during training. This could result in a potential risk of overfitting.
This is caused by those structure pairs with high TM-score, which improves the threshold of similar pairs in Eq.(\ref{eq1}).
To address this issue, we further propose a substructure sampling strategy to enhance the diversity of positive samples.
This strategy has been used in the pretraining of protein structures~\cite{hermosilla2022contrastive,DBLP:conf/iclr/ZhangXJCLD023}. However, 
directly applying the empirical sampling length or ratio in these existing methods will lead to a significant decline in performance for our method. 

Due to the sensitivity of positive samples to protein structure variations, casually sampling a substructure could make a positive sample no longer similar to the query. In this paper, we propose a novel substructure sampling method. More specifically, our substructure sampling method starts from a randomly selected amino acid within the protein, and then we traverse along the protein chain towards both ends until the desired length is attained.
To make sure our sampled substructure will not deviate from our original structure too much, we use TM-score as a guidance. 
Assume the original positive sample is denoted as $P$, and the sampled substructure is denoted as $P_s$.
We aim to satisfy the following constraint:
\begin{equation}
    TM(P, P_s)\geq \alpha.   
\end{equation}
Here $\alpha$ is set to 0.9, which means the two structures are extremely similar according to the meaning of TM-score~\cite{xu2010significant}. 
Performing on-the-fly TM-score calculation during the training process would incur significant computation cost.
Hence, we calculate the minimum sampling length for each structure to satisfy the above constraint before training.
Through this finely controlled substructure sampling strategy, we achieve an increase in the diversity of positive examples while avoiding the failure of training.
As shown in Figure~\ref{figure1}, positive and negative samples are first sampled based on the query in the training phase. 
Then, the positive substructures are sampled based on the positive sample and pre-computed substructure length.

\section{Experiment}
\subsection{Evaluation Setting}
\paragraph{Datasets} We employ three datasets in our experiment, which are SCOPe v2.07~\cite{DBLP:journals/nar/FoxBC14} , ind\_PDB~\cite{xia2022fast}, and Alphafold database~\cite{varadi2022alphafold}.
SCOPe is a database to study protein structural relationships, which has been utilized in existing works~\cite{liu2018learning,xia2022fast}. 
To ensure a fair comparison, we employ the same filtering criteria as in~\cite{xia2022fast}, and the resulting dataset contains 14,215 protein structures. 
ind\_PDB has also been utilized in~\cite{xia2022fast}, which consists of 1,900 protein structures collected from the Protein Data Bank~\cite{berman2000protein}. 
AlphaFold database contains over 200 million protein structures predicted by Alphafold 2.
Following existing work~\cite{liu2018learning,xia2022fast}, we first conduct 5-fold cross-validation on the SCOPe dataset and then evaluate the performance on the ind\_PDB dataset. AlphaFold database is mainly used for comparing memory cost.

\paragraph{Metrics}   
We employ three metrics to evaluate the accuracy: the area under the receiver operating characteristic curves (AUROC), the area under the precision-recall curves (AUPRC), and the Top-k hit ratio. 
For each query, we calculate the AUROC and AUPRC, and report the average value across all queries. 
The Top-k hit ratio is computed as $\frac{1}{N_{P_Q}}\sum_{i=1}^{N_{P_Q}}\frac{N_{\text{hit}}^{i}}{\min(k, N_{\text{nbr}}^i)}$, where $N_{P_Q}$ represents the number of query structures, $N_{\text{hit}}^i$ denotes the number of correctly identified similar structures in the top-k rankings, and $N_{\text{nbr}}^i$ indicates the total number of similar structures for the given query $P_Q$. 
Our evaluation considers the Top-k values of Top-1, Top-5, and Top-10. 
We follow the settings of existing works~\cite{liu2018learning, xia2022fast}, considering structure pairs with a TM-score larger than or equal to $0.9\cdot \max\{{\text{TM}(P_Q,P_i) |{P_i\in\mathcal{D}}}\}$ as similar pairs. 

To prove the efficiency of POSH, we also compare POSH with baselines in terms of time and memory cost.
\paragraph{Baselines}
We adopt four alignment-free methods as baselines in our experiment, which include SGM \cite{rogen2003automatic}, SSEF \cite{zotenko2006secondary}, DeepFold \cite{liu2018learning} and GraSR \cite{xia2022fast}. SGM and SSEF are non-learning methods. DeepFold and GraSR are learning-based methods. 
The results of SGM and SSEF  are obtained by running their provided scripts. 
For DeepFold, we train it on our dataset using the same settings described in the original paper. 
The results of GraSR are directly copied from its original paper for comparison since we utilize the same dataset and filtering criteria.

\subsection{Results}
During the testing phase, we aim to search protein structures similar to a given query structure from a database. 
In our experiment, the training set serves as the database, and the structures in the validation or test set serve as queries.
The ranking of structures is based on the Hamming distance between the binary vectors of protein structure pairs.

\paragraph{Accuracy on SCOPe} 
We compare the accuracy of  POSH with baselines on SCOPe dataset. 
The results are shown in Table \ref{table1}. 
We can find that POSH outperforms all other methods on all metrics except the Top-1 hit ratio. For the Top-1 hit ratio, POSH is comparable to the best baseline, and significantly outperforms other baselines.

\paragraph{Accuracy on ind\_PDB} 
We also evaluate the accuracy of POSH and baselines
on ind\_PDB.
The results are shown in Table~\ref{table2}. 
Compared with the SCOPe dataset, the accuracy of all the methods on the ind\_PDB dataset drops.
The gap can be attributed to the fact that each protein structure in the ind\_PDB dataset may consist of multiple domains, but in SCOPe each protein structure represents a single domain. 
From another perspective, ind\_PDB is a dataset more related to real-world scenarios.
The results show that POSH consistently outperforms all the other methods across all evaluation metrics on the ind\_PDB dataset.
We can find that POSH achieves about 3\%-7\% improvement on Top-k hit ratio compared to the state-of-the-art baselines.

\begin{table}[t]
\caption{Accuracy on SCOPe}
  \label{table1}
  \centering
  \scalebox{0.8}
  {
  \begin{tabular}{lccccc}
    \toprule        
    Model     &AUROC    &AUPRC    & Top-1   & Top-5   & Top-10    \\
    \midrule
    SGM       & 0.9043  & 0.4834  & 0.5633  & 0.5264  & 0.5521    \\
    SSEF      & 0.8418  & 0.0391  & 0.0830  & 0.0608  & 0.0638    \\
    DeepFold  & 0.9628  & 0.5175  & 0.5935  & 0.5680  & 0.5953    \\
    GraSR     & 0.9823  & 0.6595  & \textbf{0.7282}  & 0.7101  & 0.7400    \\
    \midrule
    POSH     & \textbf{0.9906}  & \textbf{0.6853}  & 0.7242  &   \textbf{0.7225} & \textbf{0.7592}    \\
    \bottomrule
  \end{tabular}
  }
\end{table}

\begin{table}[t]
\caption{Accuracy on ind\_PDB}
  \label{table2}
  \centering
  \scalebox{0.8}
  {
  \begin{tabular}{lccccc}
    \toprule        
    Model     &AUROC    &AUPRC& Top-1   & Top-5   & Top-10    \\
    \midrule
    SGM       & 0.8750  & 0.2563  & 0.3010  & 0.2881  & 0.3026    \\
    SSEF      & 0.8329  & 0.0323  & 0.0516  & 0.0402  & 0.0438    \\
    DeepFold  & 0.9389  & 0.3051  & 0.3358  & 0.3393  & 0.3678    \\
    GraSR     & 0.9528  & 0.4058  & 0.4558  & 0.4488  & 0.4764    \\
    \midrule
    POSH     & \textbf{0.9699}  & \textbf{0.4719}  & \textbf{0.4921}  & \textbf{0.5130}  & \textbf{0.5550}    \\
    \bottomrule
  \end{tabular}
  }
\end{table}

\begin{wrapfigure}{r}{0.5\textwidth}
  \centering
  \includegraphics[width=0.40\textwidth]{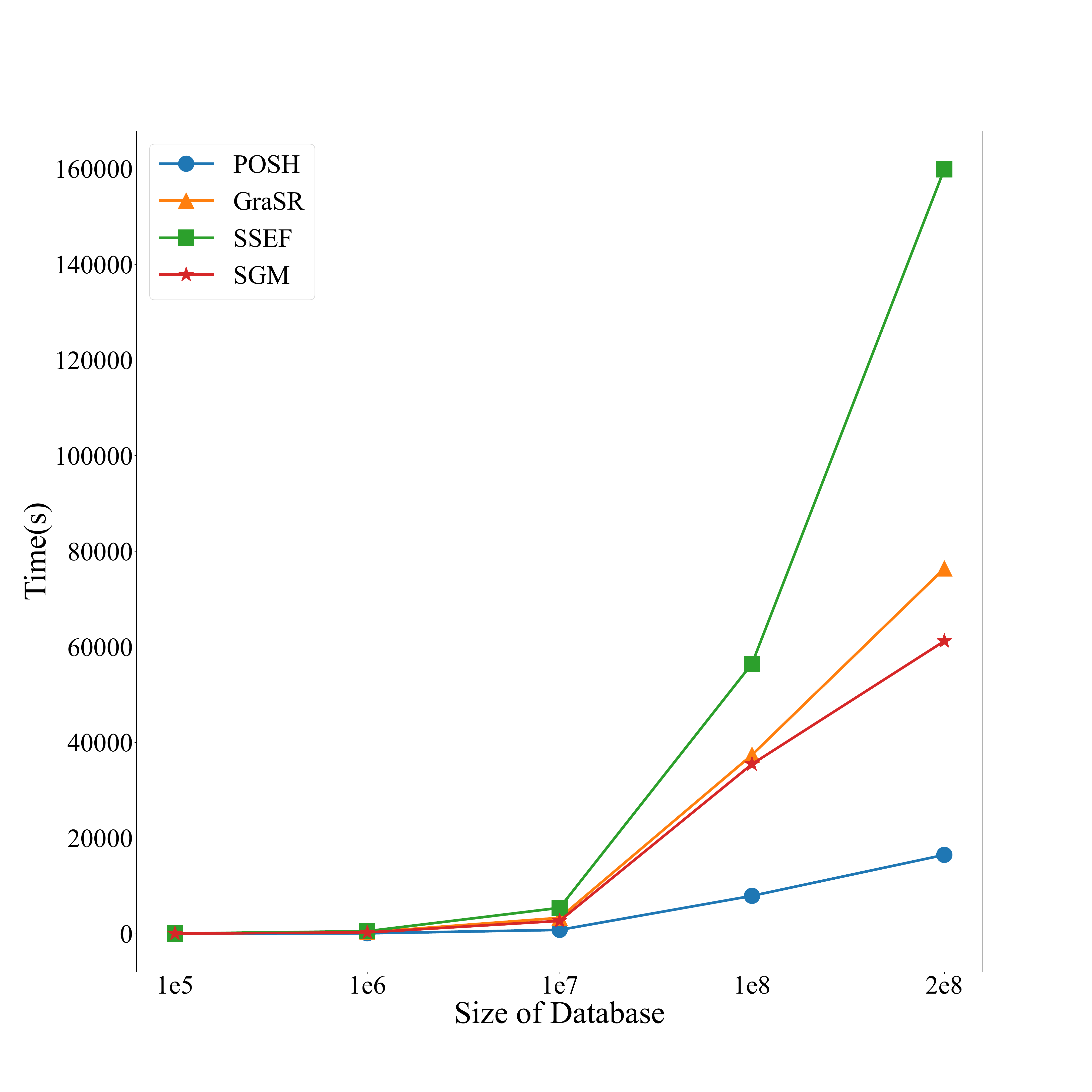}  
  \caption{Time cost of searching in databases of different sizes}  
  \label{time}
\end{wrapfigure}

\begin{table*}[t]
\caption{Comparison of memory cost}
  \label{table3}
  \centering
  \scalebox{0.8}{
  \begin{tabular}{lrr|rr}
    \toprule        
    Model      &SCOPe           &ind\_PDB        &Alphafold Database  &Compression Ratio\\
    \hline    
    TM-align   & 3952.63MB      & 4380.84MB      & 21TB         &1$\times$\\
    SGM        & 4.74MB         & 5.47MB         & 72GB         &299$\times$\\ 
    SSEF       & 286.32MB       & 324.32MB       & 1196GB       &18$\times$\\
    DeepFold   & 20.24MB        & 23.14MB        & 319GB        &67$\times$\\
    GraSR      & 20.29MB        & 23.19MB        & 320GB        &67$\times$\\
    \hline
    POSH     & \textbf{0.71MB}&\textbf{0.83MB} & \textbf{11GB}& \textbf{1955} $\times$\\    
    \bottomrule
  \end{tabular}
  }
\end{table*}

\begin{table}[t]
\caption{Ablation study}
  \label{table5}
  \centering
  \scalebox{0.8}
  {
  \begin{tabular}{ccc|ccccc}
    \toprule        
Edge Features&    Substructure Sampling & Distance Scaling    &AUROC    &AUPRC    & Top-1    & Top-5     & Top-10   \\
    \midrule
    \Checkmark&\Checkmark& \XSolidBrush  &0.9685   &0.4589   &0.4811    &0.5069     & 0.5489   \\
    \Checkmark&\XSolidBrush& \Checkmark  &0.9695   & 0.4498  & 0.4758   & 0.4960    &0.5424     \\
    \XSolidBrush&\Checkmark &\Checkmark   & 0.9651  &0.4002   &0.4237   &0.4372   &0.4826    \\
    \midrule
    \Checkmark&\Checkmark &\Checkmark       & \textbf{0.9699}  & \textbf{0.4719}  & \textbf{0.4921}  & \textbf{0.5130}  & \textbf{0.5550}    \\
    \bottomrule
  \end{tabular}
  }
\end{table}

\paragraph{Time Cost}
We show the time cost of searching in databases of different sizes for SGM, SSEF, GraSR and POSH in Figure~\ref{time}. 
We use ind\_PDB as the query dataset, and the number of protein structures in the target database increases from 100,000 to the size of the Alphafold database.
Because the search time of TM-align ranges from hours to days, we exclude it in the figure.
We have also excluded the results of DeepFold because DeepFold embeds the protein structure into the same dimension as GraSR,  which results in the same time cost.
We can find that the time cost of POSH is significantly lower than that of the other methods.
POSH is nearly four and ten times faster than SGM and SSEF, and more than four times faster than the other best method GraSR.
The advantage of POSH becomes more pronounced with the increasing database size, ensuring POSH's fast search of large-scale data.

\paragraph{Memory Cost}  
The memory cost and memory compression ratio of TM-align, POSH, and the other alignment-free methods are shown in Table~\ref{table3}. 
The process of encoding a protein structure into a vector can be seen as a form of compression, so we also calculate the compression ratio for the Alphafold database.
For TM-align, all protein structures must be stored on the computing device to calculate pairwise similarity. 
In contrast, alignment-free methods allow for memory saving by representing each protein structure as a vector. Hence, we can perform computation on the computing device with the vector representation, and then index the desired proteins on the storage device. 

From the results in Table~\ref{table3}, we can find that the alignment-free methods can dramatically reduce the memory cost, compared with the alignment-based method TM-align. Compared with existing alignment-free methods, POSH achieves a memory saving of more than six times, by up to more than 100 times. Taking Alphafold database as an example. When calculating the TM-score by utilizing TM-align directly on the original protein structure data, the resulting memory cost is 21TB, which is unmanageable. 
Even if the protein structures are represented as real-valued vectors by using some alignment-free methods, the memory cost is still high.
Specifically, the memory costs of SGM and SSEF are 72GB and 1196GB, which are 6.5 and 108.7 times larger than that of POSH, respectively.
It takes over 300GB memory cost for the other two learning-based methods DeepFold and GraSR, which is about 32 times larger than that of POSH.
The memory cost of POSH is only 11GB, which is acceptable for many real-world applications.

\subsection{Ablation Study}
We conduct ablation study on the edge updating scheme, substructure sampling strategy, and distance scaling strategy of POSH. The results are presented in Table~\ref{table5}. 
To verify the effectiveness of our edge features in capturing the structural similarity, we replace the raw edge features with those of GraSR and correspondingly remove the edge update layer.
We can find that the hand-crafted raw edge features we construct and the structure encoder we design contribute significantly to the improvement of accuracy.
We can also find that our substructure sampling strategy contributes 1\% to 2\% for Top-k hit ratio.
The results also verify the importance of adopting distance scaling to differentiate the samples with the same Hamming distance.

\section{Conclusion}
In this paper, we propose a novel method called POSH for protein structure similarity search.
POSH learns a binary vector~(hash code) representation for each protein structure, which can dramatically reduce the time and memory cost compared with real-valued vector representation based methods. 
Experimental results show that POSH can outperform other methods to achieve the best performance, in terms of accuracy, time cost and memory cost.

\bibliography{ref}
\appendix

\section{Implementation Details}
\label{implementation_details}
In our implementation, we set the value of $\gamma$ to 0.2 and $\lambda$ to 0.5. 
The temperature coefficient $\tau$ is set to 0.07. 
In each time, we sample 64 proteins: one query, one positive sample, and 62 negative samples. 
To ensure stable training, we utilize gradient accumulation with 40 accumulation steps. 
The number of layers in the structure encoder is 6. 
The code length of the binary hash code is set to 400, which is the same as that of the real-valued vectors used in existing learning-based methods~\cite{liu2018learning, xia2022fast}. 
$\rho$ is set to 0.9.
We employ the Adam optimizer with a learning rate of 0.0003 for optimization. 
Our model is trained on NVIDIA RTX A6000 GPUs, and each model is trained up to 100 epochs. 

\section{Additional Experiments}
\begin{figure}[t]
\centering
\includegraphics[scale=0.50]{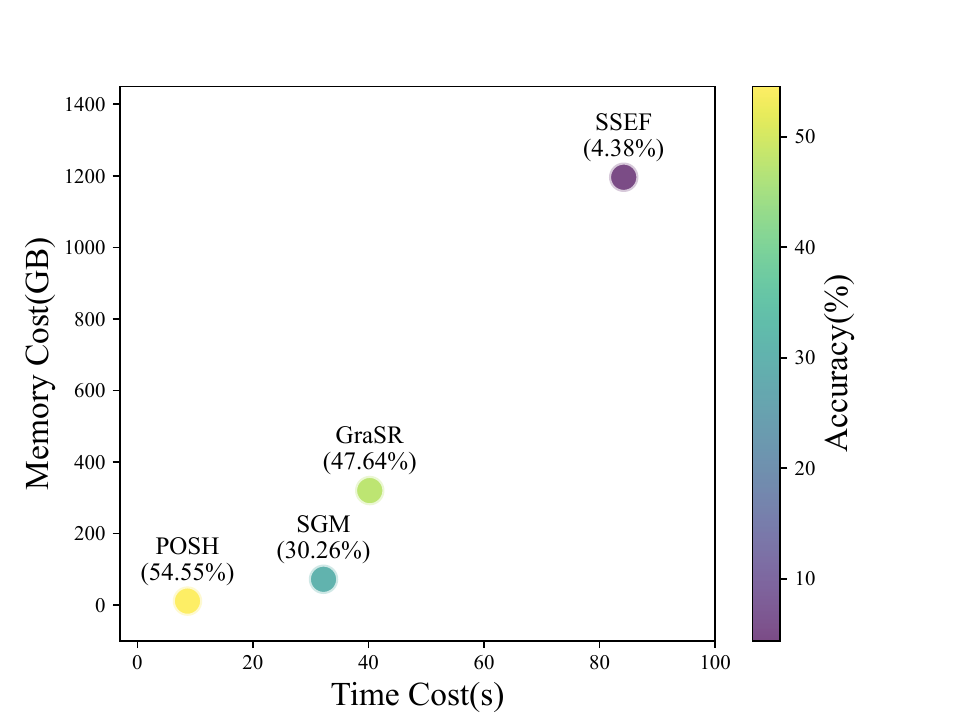}
\caption{Comprehensive performance comparison} 
\label{overall}
\end{figure}
\subsection{Comprehensive Analysis}
In Figure \ref{overall}, we provide a  comprehensive comparison of accuracy, time cost, and memory cost for SGM, SSEF, GraSR and POSH.
We can find that POSH can achieve faster speed, consume less memory, and conduct more accurate search simultaneously.
Compared with the existing most efficient method SGM, POSH achieves an accuracy improvement of 20.29\%.
Compared with the existing baseline of best accuracy~(GraSR), POSH achieves a memory saving of 32 times and speed improvement of four times.
The result shows the promising potential of POSH in PSSS.
\subsection{Code Length Experiment}
To investigate the effect of code length, we train four models with code length to be 64, 128, 256, 400, and 512 respectively.
The result is shown in Table \ref{code_length}.
We can find that with the growing code length, the accuracy of the models consistently improves. Larger code length will lead to larger time and memory cost. Hence, in real applications, we need to choose a suitable code length to achieve a good trade-off between accuracy and cost. Our POSH method provides a good choice to achieve such a trade-off. 
\begin{table}[t]
\caption{Accuracy of POSH with different code length}
  \label{code_length}
  \centering
  \scalebox{0.8}
  {
  \begin{tabular}{lccccc}
    \toprule        
    Model     &AUROC    &AUPRC& Top-1   & Top-5   & Top-10    \\
    \midrule
    POSH-64   & 0.9582  & 0.3327  & 0.3468  & 0.3752  & 0.4334    \\
    POSH-128  & 0.9641  & 0.3792  & 0.3905  & 0.4176  & 0.4696    \\
    POSH-256  & 0.9720  & 0.4356  & 0.4616  & 0.4769  & 0.5284    \\    
    POSH-400$\star$  & 0.9708  & 0.4622  & 0.4858  & 0.5049  & 0.5455    \\
    POSH-512  & 0.9663  & 0.4810  & 0.5021  & 0.5176  & 0.5665 \\
    \bottomrule
  \end{tabular}
  }
\end{table}
\section{Limitations}
\label{limitations}
In this section, we will discuss its limitations. Firstly, in the hashing layer of POSH, various hashing techniques can be explored to address the binarization problem. Future research can investigate whether there are more suitable hashing techniques specifically designed for protein structure hashing. Secondly, as the number of known protein structures increases, utilizing more protein structure data in our training set becomes possible. However, the upper limit of our model’s performance when scaling up with the
amount of data still needs to be explored. It is important to consider the challenges associated with using the original time-consuming alignment-based method for similarity calculation as the dataset grows, along with the cost of model training.




\end{document}